\documentclass[letterpaper, 10 pt, conference]{./IEEEconf}

\IEEEoverridecommandlockouts
\usepackage{amsmath}
\usepackage{amssymb,bm}
\usepackage{amsfonts}
\usepackage{pifont}
\newcommand{\cmark}{\ding{51}}
\newcommand{\xmark}{\textcolor{black!25}{\ding{55}}}
\usepackage{enumerate}
\usepackage{tabulary}
\usepackage{multirow}
\usepackage{cite}
\usepackage{lipsum} 
\usepackage{verbatim}
\usepackage{hyperref}
\usepackage{url}
\usepackage{balance}
\usepackage{graphicx}
\usepackage{epstopdf}
\usepackage[misc]{ifsym}
\usepackage{color}
\usepackage[table]{xcolor}
\usepackage{xxcolor}
\usepackage{tikz}
\usepackage{pgf}
\usepackage{pgfplots}
\usepackage{float}
\usepackage{ifthen}
\usepackage{forloop}
\usepackage{listings}
\usepackage{booktabs}
\usepackage[lined,boxed,commentsnumbered,linesnumbered]{algorithm2e}

\definecolor{codegreen}{rgb}{0,0.6,0}
\definecolor{codepurple}{rgb}{0.58,0,0.82}
\definecolor{backcolour}{rgb}{0.95,0.95,0.92}
\lstdefinestyle{terminal}{
    backgroundcolor=\color{black!5},   
    commentstyle=\color{codegreen},
    keywordstyle=\color{blue},
    numberstyle=\tiny\color{black!30},
    stringstyle=\color{codepurple},
    basicstyle=\footnotesize\ttfamily,
    breakatwhitespace=false,         
    breaklines=true,                 
    captionpos=b,                    
    keepspaces=true,                 
    numbers=left,                    
    numbersep=5pt,                  
    showspaces=false,                
    showstringspaces=false,
    showtabs=false,                  
    tabsize=2,
}
\pgfplotsset{compat=newest}
\usepgfplotslibrary{groupplots,fillbetween}

\usetikzlibrary{arrows,shadows,petri,automata,shapes.geometric}
\usetikzlibrary{positioning,fit,calc,angles,quotes,fadings,decorations.pathreplacing}
\usepackage{tikz-3dplot}
\usetikzlibrary{decorations} \usetikzlibrary{arrows.meta}
\usetikzlibrary{shapes.arrows}
\usetikzlibrary{spy}
\usetikzlibrary{shapes,backgrounds}
\usepgfplotslibrary{statistics}

\newcommand{\cameraready}[1]{{\color{black} #1}}

\SetAlCapSkip{-0.25em}
\title{\LARGE \bf
{\tt \Large aerial-autonomy-stack}---a Faster-than-real-time, Autopilot-agnostic,\\ ROS2 Framework to Simulate and Deploy Perception-based Drones
}
\author{
    Jacopo Panerati$^{\dag}$, Sina Sajjadi$^{\dag}$, Sina Soleymanpour$^{\dag}$, Varunkumar Mehta$^{\dag}$, and Iraj Mantegh$^{\dag}$
    \thanks{
        $^{\dag}$Jacopo Panerati, Sina Sajjadi, Sina Soleymanpour, Varunkumar Mehta, and Iraj Mantegh are with the National Research Council Canada;
        e-mails: {\tt \{name.lastname\}@cnrc-nrc.gc.ca}.
    }
}
\begin{document}
\maketitle
\thispagestyle{empty}
\pagestyle{empty}
\begin{abstract}
  Unmanned aerial vehicles are rapidly transforming multiple applications, from agricultural and infrastructure monitoring to logistics and defense. 
  Introducing greater autonomy to these systems can simultaneously make them more effective as well as reliable.  
  Thus, the ability to rapidly engineer and deploy autonomous aerial systems has become of strategic importance. 
  In the 2010s, a combination of high-performance compute, data, and open-source software led to the current deep learning and AI boom, unlocking decades of prior theoretical work. 
  Robotics is on the cusp of a similar transformation. 
  However, physical AI faces unique hurdles, often combined under the umbrella term ``simulation-to-reality gap''. 
  These span from modeling shortcomings to the complexity of vertically integrating the highly heterogeneous hardware and software systems typically found in field robots. 
  To address the latter, we introduce \texttt{aerial-autonomy-stack}, an open-source, end-to-end framework designed to streamline the pipeline from (GPU-accelerated) perception to (flight controller-based) action. 
  Our stack allows the development of aerial autonomy using ROS2 and provides a common interface for two of the most popular autopilots: PX4 and ArduPilot. 
  We show that it supports over 20$\times$ faster-than-real-time, end-to-end simulation of a complete development and deployment stack---including edge compute and networking---significantly compressing the build-test-release cycle of perception-based autonomy.
\end{abstract}
\vspace{1.25em}
Open-source repository: \href{https://github.com/JacopoPan/aerial-autonomy-stack}{\texttt{aerial-autonomy-stack}} 
\vspace{0.25em}
\section{Introduction}
\label{sec:intro}

The research domain of aerial robotics, encompassing compact consumer drones as well as sophisticated uncrewed aerial systems (UAS), is witnessing explosive growth. 
Over the last decade, platforms have evolved from simple aerial photography tools into high-performance machines capable of high-speed racing and precision agriculture. 
Modern applications range from search-and-rescue missions to automated power line inspections.

All of the latest developments have been driven by a combination of priorities, pioneered by industry, academia, or defense. 
Commercial companies are deploying autonomous delivery networks, aerial survey fleets, and drone shows with thousands of robots. 
The academic community has pushed the envelope of high-speed drone racing beyond human performance~\cite{uzh-racing,tii-racing}.
Recent geopolitical events highlighted the versatility of drones in security beyond traditional intelligence, surveillance, and reconnaissance roles. 
Autonomy can now serve as a crucial force multiplier, enabling effective operation within electronically degraded environments and contested airspace. 
Hence, the ability to engineer and deploy robust aerial autonomy represents a vital strategic capability~\cite{ai-uav,floreano,cloud}.

Despite this demand, the development landscape for aerial robotics remains significantly fragmented~\cite{keynote-oss}.
Simulation tools vary widely and tend to be tailored to specific applications and niches~\cite{sim-survey}.
A strong ecosystem of open-source autopilots exists, yet the associated development tools typically focus on real-time, single-vehicle simulation, using standard sensors like barometers and IMUs, for low-level debugging. 
In contrast, advanced perception tools---such as YOLO for object detection and SLAM algorithms---are often developed independently from the control stack. 
This disconnect results in a scarcity of fully integrated solutions---at odds with the aerospace industry's long-standing reliance on vertical integration for reliability.

\begin{figure}
    \centering
    \includegraphics[
                width=\columnwidth,
trim={30.0cm 0.0cm 0.0cm 3.0cm},clip
    ]{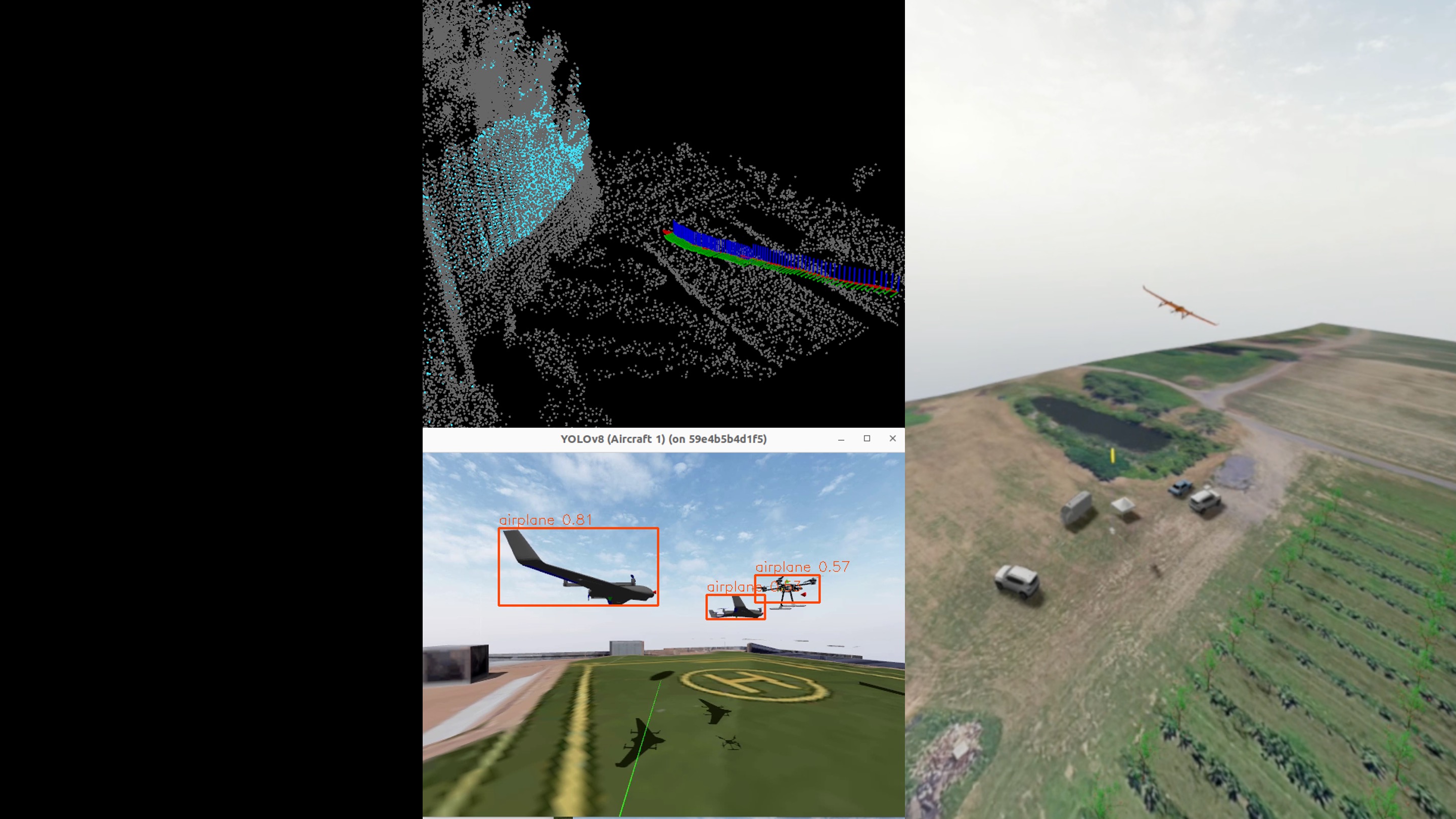}
    \vspace{-1em}
    \caption{
    Gazebo Sim scene rendering (right), FPV camera view with YOLO overlays (bottom left), and RViz LiDAR point cloud visualization (top left).
    }
    \label{fig:frontpage}
\end{figure}

\begin{table*}
    \centering
    \caption{
    Comparison of Frameworks to Develop, Simulate, and Validate Aerial Autonomy
    }
    \hspace{-1.8em}
\scalebox{0.721}{

\setlength{\aboverulesep}{0pt}
\setlength{\belowrulesep}{0pt}
\renewcommand{\arraystretch}{1.25}

\begin{tabular}{
>{\centering\arraybackslash}m{3.75cm}
>{\centering\arraybackslash}m{1.85cm} 
>{\centering\arraybackslash}m{2.05cm} 
>{\centering\arraybackslash}m{1.65cm} 
>{\centering\arraybackslash}m{1.45cm} 
>{\centering\arraybackslash}m{1.45cm} 
>{\centering\arraybackslash}m{1.0cm} 
>{\centering\arraybackslash}m{1.0cm} 
>{\centering\arraybackslash}m{1.35cm} 
>{\centering\arraybackslash}m{1.35cm} 
>{\centering\arraybackslash}m{1.35cm} 
>{\centering\arraybackslash}m{1.40cm} 
}
    \toprule

    {\bfseries Software Stack}
    & {\bfseries Autopilot Support}
    & {\bfseries Middleware Integrations}
    & {\bfseries Target/Edge Architectures}
    & {\bfseries Types of Vehicles}
    & {\bfseries Multi-agent Support}
    & {\bfseries Camera Support}
    & {\bfseries LiDAR Support}
    & {\bfseries SITL Simulation}
    & {\bfseries HITL Simulation}
    & {\bfseries Docker Containers}
    & {\bfseries Simulation Backend(s)}
    \tabularnewline

    \cmidrule(lr){1-1}
    \cmidrule(lr){2-4}
    \cmidrule(lr){5-6}
    \cmidrule(lr){7-8}
    \cmidrule(lr){9-10}
    \cmidrule(lr){11-11}
    \cmidrule(lr){12-12}

{\tt \footnotesize Aerialist}~\cite{aerialist} &
    PX4 & 
    ROS1, MAVLink & 
    x86 &
    Multicopters &
    \xmark & \xmark & \xmark & \cmark & \xmark & \cmark &
    Gazebo, jMAVSim \tabularnewline
    \cmidrule(lr){1-12}
\rowcolor{black!10}
    \color{black} {\tt \footnotesize aerial-autonomy-stack}  
    [this work] &
    \color{black} PX4, ArduPilot & 
    \color{black} ROS2, \textmu DDS, MAVROS, Zenoh  & 
    \color{black} x86, Jetson Orin & 
    \color{black} Multicopters, VTOLs & 
    \color{black} \cmark & \color{black} \cmark & \color{black} \cmark & \color{black} \cmark & \color{black} \cmark & \color{black} \cmark &
    \color{black} Gazebo \tabularnewline
    \cmidrule(lr){1-12}
{\tt \footnotesize aerostack2}~\cite{aerostack2} &
    PX4, Tello, DJI, Crazyflie & 
    ROS2, MAVLink, BehaviorTree & 
    x86, Jetson Xavier &
    Multicopters &
    \cmark & \cmark & \cmark & \cmark & \cmark & \cmark &
    Gazebo, DJI HITL Simulator \tabularnewline
    \cmidrule(lr){1-12}
{\tt \footnotesize agilicious}~\cite{agilicious} &
    agiNuttx, Betaflight & 
    ROS1, agilib, agiros & 
    x86, Jetson TX2, Xavier &
    Quadcopters &
    \xmark & \cmark & \xmark & \cmark & \cmark & \cmark &
    BEM, Flightmare \tabularnewline
    \cmidrule(lr){1-12}
{\tt \footnotesize crazyswarm2}~\cite{crazyswarm} &
    Crazyflie & 
    ROS1, ROS2 & 
    x86, STM32 &
    Crazyflie &
    \cmark & \cmark & \cmark  & \cmark & \xmark & \cmark &
    Custom \tabularnewline
    \cmidrule(lr){1-12}
{\tt \footnotesize grvc-ual}~\cite{grvc} &
    PX4, ArduPilot, DJI, Crazyflie & 
    ROS1, MAVROS, \hspace{2em} DJI SDK & 
    x86 &
    Multicopters &
    \cmark & \cmark & \xmark & \cmark & \cmark & \xmark &
    Gazebo, AirSim, DJI HITL Sim.\tabularnewline
    \cmidrule(lr){1-12}
{\tt \footnotesize kr\_autonomous\_flight}~\cite{upenn-kr-autonomous-flight} &
    PX4 & 
    ROS1, MAVROS & 
    x86, NUC &
    Multicopters &
    \xmark & \cmark & \cmark & \cmark & \xmark & \cmark &
    Gazebo \tabularnewline \cmidrule(lr){1-12}
{\tt \footnotesize mrs\_uav\_system}~\cite{mrs-uav} &
    PX4, Tello & 
    ROS1, ROS2, MAVROS & 
    x86, arm64 &
    Multicopters &
    \cmark & \cmark & \cmark & \cmark & \cmark & \cmark &
    Gazebo, FlightForge \tabularnewline
    \cmidrule(lr){1-12}
{\tt \footnotesize XTDrone}~\cite{xtdrone} &
    PX4 & 
    ROS1, MAVROS & 
    x86 &
    Multicopters, VTOLs, etc. &
    \cmark & \cmark & \cmark & \cmark & \xmark & \xmark &
    Gazebo \tabularnewline

    \bottomrule
\end{tabular}
}
\vspace{-1em}

     \label{tab:comparison}
    \vspace{-1.5em}
\end{table*}

With \texttt{aerial-autonomy-stack} (Figure~\ref{fig:frontpage}), we aim to close this gap by introducing a unified, open-source~\cite{code-impact} research framework for the development and deployment of aerial robots. 
The contributions of this work include:

\begin{itemize}
    \item a unified development environment---we provide a development playground that supports perception-enabled (with RGB cameras and 3D LiDARs) autonomy for both the PX4 and ArduPilot ecosystems, streamlining cross-platform research;
    \item a standardized autopilot interface---our stack introduces an autopilot-agnostic ROS2 action-based interface for PX4 and ArduPilot, to implement autonomy using modern software design principles;
    \item high-performance simulation---we demonstrate faster-than-real-time, multi-vehicle, software-in-the-loop capabilities, performing end-to-end simulation that includes RGB camera feeds, LiDAR odometry, and onboard GPU-accelerated YOLO inference;
    \item a containerized architecture---we leverage Docker containers to achieve three critical goals: 
    \begin{itemize} 
        \item accurately emulate the complex networking topologies of multi-agent systems;
        \item enable multi-target compilation, allowing the same source code to be used for both simulation and physical deployment;
        \item facilitate distributed hardware-in-the-loop simulation, incorporating NVIDIA Jetson modules and RF communication links, to validate onboard compute and wireless configurations. 
    \end{itemize} 
\end{itemize} \section{Related Work}
\label{sec:related}

The existing landscape\footnote{
\href{https://github.com/ROS-Aerial/aerial_robotic_landscape}{\texttt{https://github.com/ROS-Aerial/}}
} of open-source autopilots provides a robust foundation for flight control, but typically does not include out-of-the-box high-level autonomous capabilities. 
PX4 is a standard choice for academic research, commonly paired with Gazebo for software-in-the-loop (SITL) simulation, although it currently lacks native provisions for hardware-in-the-loop (HITL) validation and advanced perception.
ArduPilot commands significant popularity among hobbyists and commercial integrators, offering similar SITL features but facing comparable limitations regarding integrated perception-based autonomy. 

Meanwhile, Betaflight prevails in the drone racing niche but offers limited SITL support and minimal autonomy features. 
In most existing frameworks, perception modules are stacked in ad hoc fashion alongside the simulator, adding friction to deployment and reproducibility.

Aerial simulators are equally important, yet fragmented; for a comprehensive review, we refer to~\cite{sim-survey}. 
Researchers prioritizing reinforcement learning often utilize simple~\cite{gym-pybullet-drones} or high-performance~\cite{aerialgym} engines, which can offer massive data throughput and parallelization but lack the autopilot SITL integrations that are crucial for deployment. 
Similarly, high-fidelity rendering engines such as Flightmare~\cite{flightmare}, AirSim~\cite{airsim} provide the visual richness required for vision-based navigation but impose high computational overhead and are not natively integrated with the multi-target architecture software required by edge devices. 
Other simulators address specific niches, e.g. RotorPy~\cite{rotorpy} for aerodynamic fidelity.
However, few solutions merge these capabilities into a single workflow, from simulation to the physical edge.

Among the software stacks attempting to unify these components (see Table~\ref{tab:comparison}), Aerostack2~\cite{aerostack1,aerostack2} offers a ROS2 framework for multi-robot autonomy compatible with several autopilots. It relies on a plugin architecture that can add flexibility but also introduce integration overhead. 
The MRS System~\cite{mrs-uav}, originally built on a legacy ROS architecture, has several demonstrated successes in the field deployment of multi-robot systems based on PX4.

Projects such as XTDrone~\cite{xtdrone} address multi-vehicle simulation across diverse platforms but function primarily as a simulation tool rather than a deployable, containerized stack.
UAL~\cite{grvc} provides a hardware abstraction layer for multiple autopilots but it is still based on a ROS1 framework~\cite{gesu}.

Newer frameworks such as Agilicious~\cite{agilicious} and Crazyswarm2~\cite{crazyswarm} are highly specialized for agile maneuvers and indoor nano-drone swarms, respectively, and are less applicable to general-purpose and outdoor autonomy. 
Aerialist~\cite{aerialist} focuses on automated test generation and log analysis rather than providing a deployment-time solution. 

Our stack uniquely builds upon these precedents, offering a vertically integrated alternative that simultaneously addresses the multi-agent and multi-autopilot flexibility of~\cite{aerostack2} and the perception-heavy focus of~\cite{upenn-kr-autonomous-flight}, all while supporting ROS2 and vertical take-off and landing (VTOL) platforms.
\cameraready{Furthermore, support for \emph{(i)} both ArduPilot and PX4, \emph{(ii)} Jetson-in-the-loop simulation, and \emph{(iii)} ``code parity'' between simulation and deployment on Jetson edge devices make our stack especially suited for the development of new real-world applications with minimal or no software re-work.}

\begin{figure*}[ht]
    \centering
    \begin{tikzpicture} 
    \begin{axis}[
        width = 1.05\textwidth,
        height = 7.0cm,
        xmin=0,
        xmax=99,
        ymin=0,
        ymax=90,   
        axis line style={draw=none},
        tick style={draw=none},
        xticklabel=\empty,
        yticklabel=\empty,
        clip=false,
]

\draw[black] 
    (axis cs:0, -5) rectangle (axis cs:99, 93)
    node[anchor=south west] at (axis cs:0, -5) {\ttfamily \scriptsize \bfseries amd64 host};

\node[anchor=center,draw, align=center] (models) at (axis cs:10, 80) {\ttfamily \scriptsize aircraft\_models/};
    \node[anchor=center,draw, align=center] (worlds) at (axis cs:30, 80) {\ttfamily \scriptsize simulation\_worlds/};
    \node[anchor=center,draw, text width=2.05cm, align=center] (sitl) at (axis cs:10, 60) {\scriptsize N $\times$ {\ttfamily PX4} or \\[-1ex] {\ttfamily ArduPilot SITL}};
    \node[anchor=center,draw, text width=1cm, align=center] (gz) at (axis cs:35, 60) {\ttfamily \scriptsize Gazebo\\[-1ex] Sim};
\node[draw=blue, dashed, fit=(models) (worlds) (sitl) (gz), 
    label={[anchor=south west]north west:{\ttfamily \scriptsize \color{blue} simulation-container}}] (sim_box) {};

\node[anchor=center,draw, align=center] (proxy) at (axis cs:10, 30) {\ttfamily \scriptsize mavlink-router};
    \node[anchor=center,draw, align=center] (gnd) at (axis cs:10, 17.5) {\ttfamily \scriptsize ground\_system};
    \node[anchor=center,draw, align=center] (qgc) at (axis cs:30, 30) {\ttfamily \scriptsize QGroundControl};
    \node[anchor=center,draw, align=center] (zgnd) at (axis cs:30, 10) {\ttfamily \scriptsize zenoh-bridge};
\node[draw=blue, dashed, fit=(gnd) (qgc) (zgnd) (proxy),
    label={[anchor=south west]north west:{
}}] (gnd_box) {};

\node[anchor=center,draw, align=center] (zair) at (axis cs:60, 10) {\ttfamily \scriptsize zenoh-bridge};
    \node[anchor=center,draw, align=center] (ss) at (axis cs:90, 10) {\ttfamily \scriptsize state\_sharing};
    \node[anchor=center,draw, align=center] (yolo) at (axis cs:60, 80) {\ttfamily \scriptsize yolo\_py};
    \node[anchor=center,draw, align=center] (kiss) at (axis cs:60, 60) {\ttfamily \scriptsize kiss\_icp};
    \node[anchor=center,draw, text width=1.5cm, align=center] (dds) at (axis cs:60, 40) {\scriptsize {\ttfamily uxrce\_dds}\\[-1ex] or {\ttfamily MAVROS}};
    \node[anchor=center,draw, align=center] (offboard) at (axis cs:80, 70) {\ttfamily \scriptsize offboard\_control};
    \node[anchor=center,draw, align=center] (mission) at (axis cs:92.5, 80) {\ttfamily \scriptsize mission};
    \node[anchor=center,draw, align=center] (ap) at (axis cs:87.5, 35) {\ttfamily \scriptsize autopilot\_interface};
\node[draw=blue, dashed, fit=(zair) (ss) (yolo) (kiss) (dds) (offboard) (mission) (ap),
    label={[anchor=south west]north west:{\scriptsize N $\times$ {\ttfamily \color{blue} aircraft-container\_[1..N]}}}] (air_box) {};

    \draw[-latex] (models) to[out=320, in=150] (gz);
    \draw[-latex] (worlds) to[out=270, in=90] (gz);
    \draw[latex-latex] (sitl) to[out=0, in=180] node[pos=0.5, rounded corners=3pt,fill=black!10,text width=1.75cm,align=center,inner sep=2pt,] {\tiny {\ttfamily gz\_bridge} or\\[-3ex] {\ttfamily ardupilot\_gazebo}} (gz);

    \draw[latex-latex,draw=teal,line width=0.75pt] (sitl) to[out=330, in=180] node[pos=0.8,text=teal, fill=white,draw,text width=1.2cm,align=center,inner sep=2pt,] {\ttfamily \tiny UDP\\[-3ex] [SIM\_SUBNET]} (dds);
    \draw[latex-latex,draw=teal,line width=0.75pt] (sitl) to[out=210, in=180] node[pos=0.31,text=teal, fill=white,draw,text width=1.2cm,align=center,inner sep=2pt,] {\ttfamily \tiny MAVLink\\[-3ex] [SIM\_SUBNET]} (proxy);
\draw[latex-latex] (proxy) to[out=330, in=180] (qgc);
    \draw[-latex] (proxy) to[out=270, in=90] (gnd);

    \draw[-latex] (gnd) to[out=300, in=180] node[pos=0.5, rounded corners=3pt,fill=black!10,,align=center,inner sep=2pt,] {\ttfamily \tiny /tracks} (zgnd);
    \draw[latex-latex,draw=red,line width=0.75pt,densely dashed] (zgnd) to[out=0, in=180] node[pos=0.5,text=red,fill=white,draw,text width=1.2cm,align=center,inner sep=2pt,] {\ttfamily \tiny TCP\\[-3ex] [AIR\_SUBNET]} (zair);
    \draw[latex-] (zair) to[out=0, in=180] node[pos=0.5, rounded corners=3pt,fill=black!10,,align=center,inner sep=2pt,] {\ttfamily \tiny /state\_drone\_[1..N]} (ss);

    \draw[-latex,draw=teal,line width=0.75pt] (gz) to[out=10, in=180] node[pos=0.45,text=teal, fill=white,draw,text width=1.4cm,align=center,inner sep=2pt,] {\ttfamily \tiny gz\_gst\_bridge\\[-3ex] [SIM\_SUBNET]} (yolo);
    \draw[-latex,draw=teal,line width=0.75pt] (gz) to[out=-10, in=180] node[pos=0.45, text=teal,fill=white,draw,text width=1.4cm,align=center,inner sep=2pt,] {\ttfamily \tiny /lidar\_points\\[-3ex] [SIM\_SUBNET]} (kiss);

    \draw[latex-latex] (dds) to[out=10, in=180] (ap);
    \draw[-latex] (dds) to[out=-10, in=170] (ss);

    \draw[-latex] (yolo) to[out=0, in=180] node[pos=0.5, rounded corners=3pt,fill=black!10,,align=center,inner sep=2pt,] {\ttfamily \tiny /detections} (offboard);
\draw[-latex] (offboard) to[out=270, in=140] node[pos=0.33, rounded corners=3pt,fill=black!10,,align=center,inner sep=2pt,] {\ttfamily \tiny /reference} (ap);
    \draw[-latex] (mission) to[out=270, in=60] node[pos=0.4, rounded corners=3pt,fill=black!10,,text width=1.1cm,align=center,inner sep=2pt,] {\ttfamily \tiny ros2\\[-3ex] action/srv} (ap);

\node[draw=none, dashed, fit=(gnd) (qgc) (zgnd),
    label={[anchor=south west]north west:{\ttfamily \scriptsize \color{blue} ground-container}}] (gnd_box) {};

    \end{axis}
\end{tikzpicture}     \caption{
    Block diagram of the software-in-the-loop simulation architecture comprising of the \emph{simulation-image}, \emph{ground-image}, and \emph{aircraft-image} containers.
}
    \label{fig:sitl}
    \vspace{-1.5em}
\end{figure*}
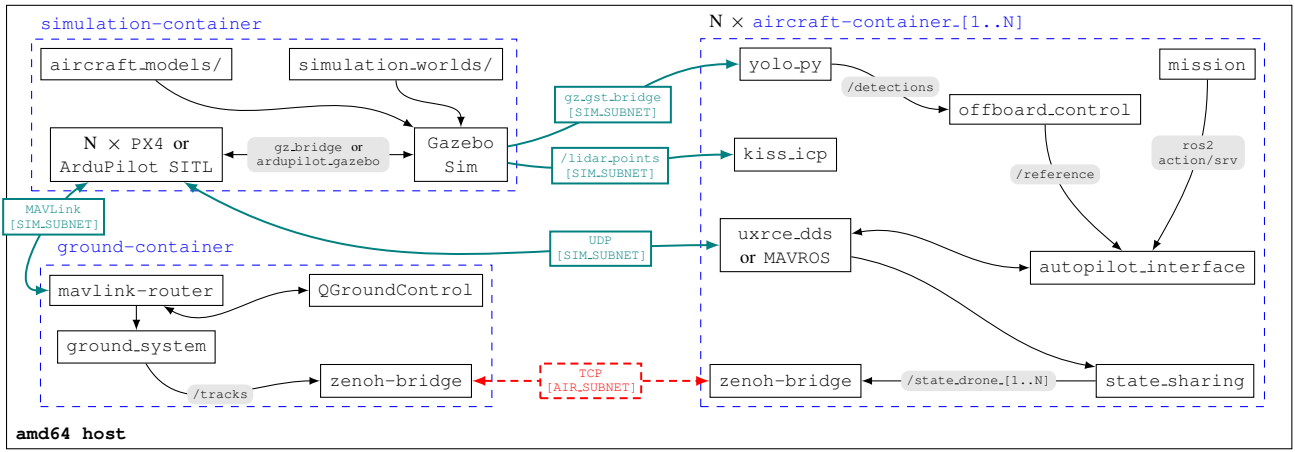
 \section{Rationale}
\label{sec:rationale}

The ``sim2real gap''~\cite{sim2realgap,remote-sim2real} is an umbrella term used to capture the many discrepancies---including in sensing and dynamic models---between simulations and the field deployment of robots.
It is often blamed for stifling progress in physical AI and robotics.
In the context of aerial robotics, sim2real research has typically focused on mitigating the errors arising from aerodynamic phenomena---such as drag, ground effect, and rotor downwash---that require non-linear and computationally expensive modeling~\cite{neural-lander}. 
However, we posit that the sim2real gap also has a system engineering aspect that is frequently underestimated. 
It is not merely a failure of physics modeling, but also a failure of integration, as significant performance degradation and information loss occur at the interfaces between disparate software and hardware components.

The \texttt{aerial-autonomy-stack} helps mitigate the system engineering aspects of the sim2real gap.
Its software architecture follows two cardinal principles: \emph{(i)} simplicity and  \emph{(ii)} end-to-endness. 
The first is paramount to maximize reliability and minimize failure modes---as code complexity correlates with bug density and maintenance costs~\cite{lean}---which is crucial for aerospace and defense applications. 
The second principle allows for cross-layer optimization along a single, vertically integrated stack---which leads to more effective compute power usage, lower friction in field deployments, and a reduced logistical footprint.
End-to-endness also enables to run integration tests directly within CI/CD pipelines, greatly expediting the development cycle.
 \section{Aerial Autonomy Stack} \label{sec:aas}

This project aims to accelerate the iterative cycle of ideation, development, simulation, testing, and deployment of autonomous drones. 
We do so by unifying these stages into a single cohesive workflow (Figure~\ref{fig:sitl} and Figure~\ref{fig:hitl}) that can drastically reduce the time-to-market and experimental risk associated with creating novel aerial robotics.

\subsection{Third-party Integrations} 

Integrating the necessary open-source tools requires rigorous version alignment and dependency management.
We consolidate all resources into a single stack to resolve the technical friction often encountered in end-to-end systems (e.g., when combining flight control firmware with AI frameworks).

\subsubsection{Simulator} We select Gazebo Sim (Harmonic)\footnote{\url{https://github.com/gazebosim}} as the core engine due to its modern architecture and modular design. 
It provides essential features such as clock synchronization with ROS2 and configurable sensor plugins for cameras and GPU-accelerated LiDARs. 
Crucially, the engine supports headless operation for CI/CD pipelines, pausing, and faster-than-real-time (FTRT) stepping. 
For its physics engine, it uses ODE, while rendering leverages Ogre2 and OpenGL, optimized for NVIDIA hardware acceleration.

\subsubsection{Autopilot Firmware} We support the two dominant open-source flight stacks to ensure broad applicability across research and industry.

\paragraph{PX4-Autopilot} PX4\footnote{\url{https://github.com/PX4/PX4-Autopilot}} is favored for its modular architecture and widespread adoption in academic research. 
We utilize the native gz-bridge interface to let the PX4 SITL target interact with Gazebo Sim. 
Our stack currently supports PX4's standard quadcopter and VTOL configurations, yet the firmware is also compatible with fixed-wing planes and, experimentally, with rovers, airships, helicopters, and more. 

\paragraph{ArduPilot and ardupilot-gazebo} ArduPilot\footnote{\url{https://github.com/ArduPilot/ardupilot}} is integrated for its proven reliability, success in commercial applications, and adoption by the hobbyist community. 
We utilize the ardupilot-gazebo\footnote{\url{https://github.com/ArduPilot/ardupilot_gazebo}} plugin to bridge the firmware SITL target with Gazebo Sim.
As for PX4, we support quadcopter and quadplane (VTOL) configurations but the ArduPilot ecosystem includes several more vehicle types, including blimps, rovers, and submarines.

\subsubsection{Middleware} A robust middleware layer is essential to enable inter-device and inter-process communication. The landscape of robotics middleware is often fragmented; we select a compact but comprehensive suite of standard tools to ensure interoperability.

\paragraph{ROS2} ROS2\footnote{\url{https://github.com/ros2}} Humble serves as the primary application layer, handling nodes and message orchestration. It acts as the backbone for high-level decision-making, connecting the autopilot interfaces with the perception and mission planning modules. 

\paragraph{DDS} Micro-XRCE-DDS\footnote{\url{https://github.com/eProsima/Micro-XRCE-DDS}} is used to handle low-level, high-frequency communication, optimizing throughput for constrained networks. 
In our stack, it serves as the primary bridge between the PX4 uORB internal messaging and the ROS2 application layer. 

\paragraph{MAVROS} MAVROS\footnote{\url{https://github.com/mavlink/mavros}} also acts as an autopilot-to-application bridge, translating MAVLink streams into ROS2 topics and services for a companion computer. 
While ArduPilot offers experimental DDS support, our stack also includes MAVROS to leverage its stabler legacy interface. 

\paragraph{GStreamer} GStreamer\footnote{\url{https://github.com/GStreamer/gstreamer}} is used for low-latency video streaming from both simulated and real-life cameras to the computer vision node. 
We implement a custom Gazebo-GStreamer bridge to capture the virtual camera buffer and encode it with hardware acceleration. 
This mimics the input for the decoding pipeline used for physical CSI cameras (e.g., IMX219), allowing the exact same perception code to run in both simulation and deployment, without modification.

\paragraph{Zenoh} We integrate Zenoh\footnote{
\href{https://github.com/eclipse-zenoh/zenoh-plugin-ros2dds}{\texttt{https://github.com/eclipse-zenoh}}
} to enable efficient peer-to-peer bridging of isolated ROS2 domains. This allows each edge device (drone) to maintain its own local namespace while selectively replicating data across unreliable wireless links often found in field operations. 

\subsubsection{Ground Segment} For mission planning and monitoring, we include both graphical and command line tools.

\paragraph{QGroundControl} We utilize QGC\footnote{\url{https://github.com/mavlink/qgroundcontrol}} for mission planning, parameter customization, and real-time flight monitoring in simulation, just as we would for real-world flights.

\paragraph{MAVLink} We utilize a custom MAVLink acquisition and routing node---bridged with Zenoh---to encapsulate select telemetry data, ensuring they are shared between the ground station and all vehicles controlled by our stack. 
We leverage its C-library implementation\footnote{\url{https://github.com/mavlink/c_library_v2}}.

\subsubsection{Perception} Autonomy requires robust interpretation of environmental data from high-bandwidth sensors.

\paragraph{YOLO} We integrate Ultralytics YOLO\footnote{\url{https://github.com/ultralytics/ultralytics}} \cite{yolo-review} for real-time object detection. We default to the nano version of the model to prioritize inference speed. 
We use it to process RGB camera streams to label objects and publish bounding boxes (using ROS2) for visual servoing. 

\paragraph{ONNX Runtime} The YOLO models are exported to the Open Neural Network Exchange\footnote{\url{https://github.com/microsoft/onnxruntime}} format to ensure portability across different GPU-accelerated inference backends (i.e., TensorRT on Jetson/deployment and CUDA on desktop/simulation).

\paragraph{KISS-ICP} For LiDAR-based odometry, we integrate KISS-ICP\footnote{\url{https://github.com/PRBonn/kiss-icp}} \cite{kiss-icp} due to its speed and minimal dependencies. 
It provides a robust ego-motion baseline estimation without the computational overhead of full optimization-based SLAM frameworks. 

\subsubsection{Analysis} Finally, we include a set of tools for post-mission analysis, debugging, and performance validation.

\paragraph{Flight Review} A web-based tool for visualizing uLog data, essential for probing into PX4 estimator states and, for real flights, vibration analysis\footnote{\url{https://github.com/PX4/flight_review}}.

\paragraph{MAVExplorer, part of MAVProxy} Provides powerful command-line log analysis for ArduPilot binary logs, supporting custom graphing and data introspection\footnote{\url{https://github.com/ArduPilot/MAVProxy}}.

\paragraph{PlotJuggler} A fast, intuitive tool for visualizing ROS2 bag files and time-series data, including PX4 logs\footnote{\url{https://github.com/facontidavide/PlotJuggler}}.

\begin{figure}
    \centering
    \includegraphics[width=\columnwidth]{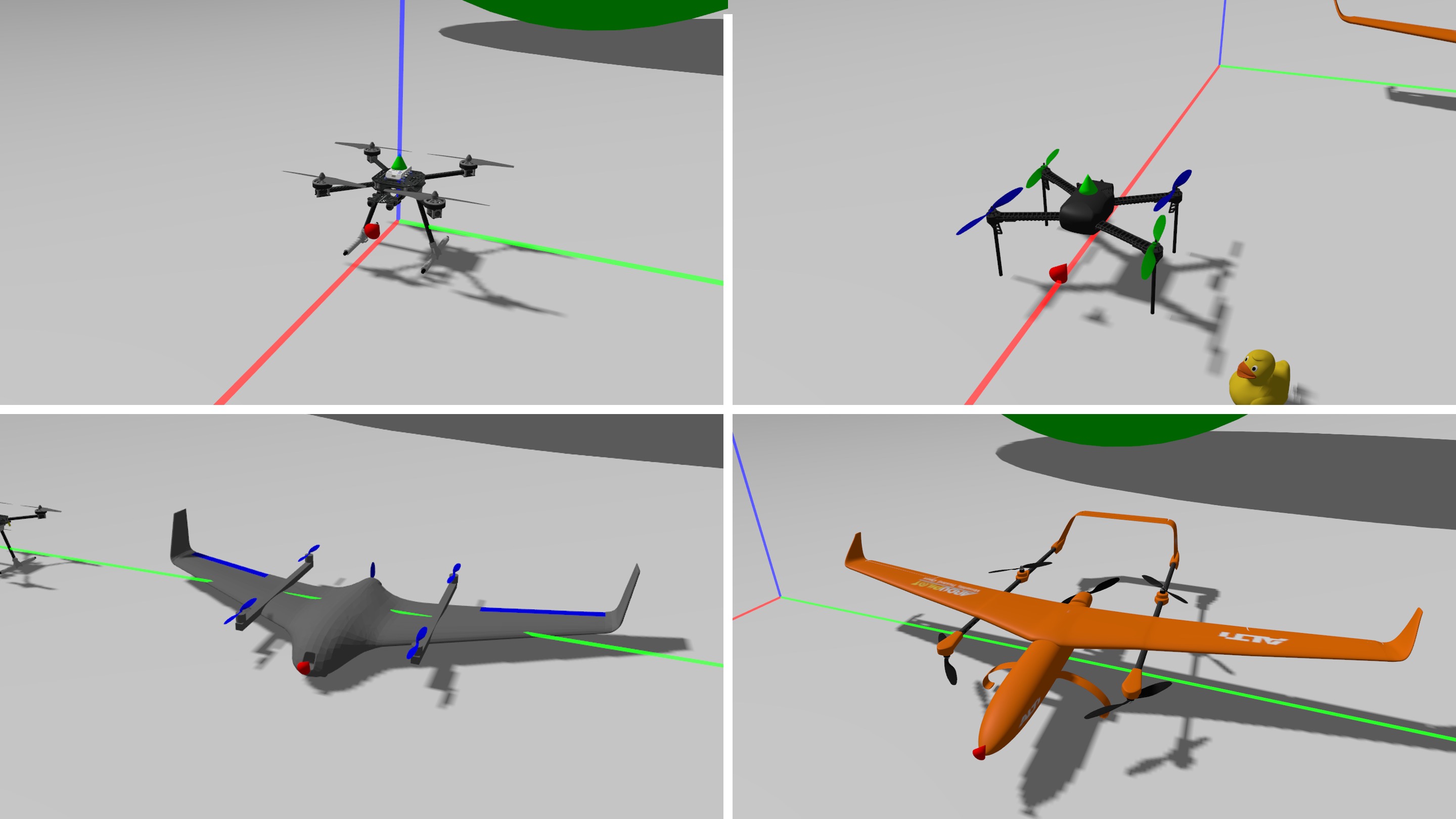}
    \vspace{-1em}
    \caption{
    Vehicle models: \emph{Holybro X500v2} (top left); \emph{3DR Iris} (top right); \emph{Standard VTOL} (bottom left); and \emph{ALTI Transition} (bottom right).
    }
    \label{fig:vehicles}
\end{figure}

\begin{figure}
    \centering
    \includegraphics[width=\columnwidth]{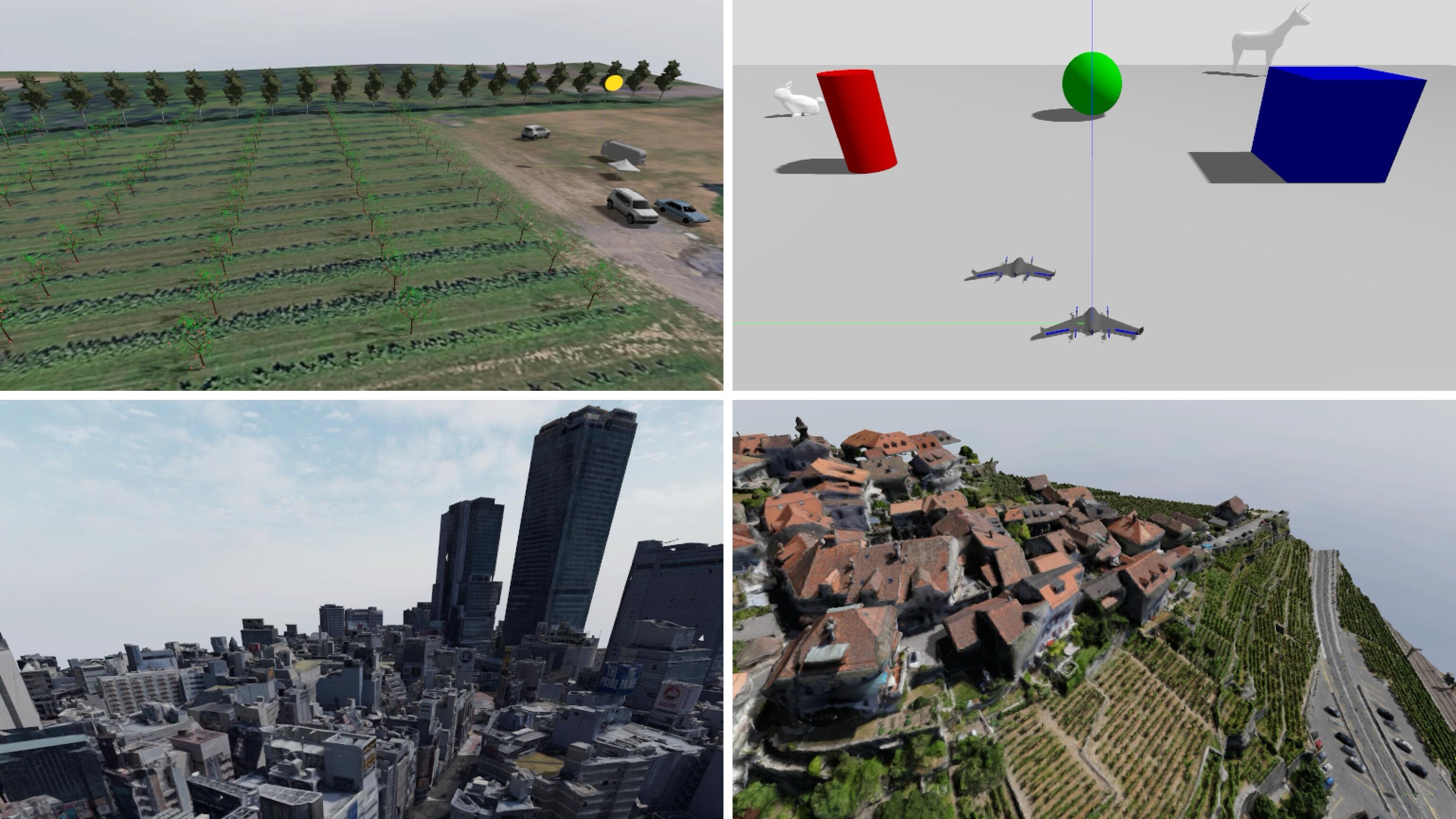}
    \vspace{-1em}
    \caption{
    3D world models: \emph{Plain} (top left); \emph{Empty} (top right); \emph{City} (bottom left); and \emph{Mountain} (bottom right).
    }
    \label{fig:worlds}
\end{figure}

\subsection{Models} 

The \texttt{aerial-autonomy-stack} includes a library of vehicles (Figure~\ref{fig:vehicles}) and simulation worlds (Figure~\ref{fig:worlds}), covering a wide spectrum of aerial applications.

\subsubsection{Aerial Vehicles} We support four SDF models provided by the PX4 and ArduPilot SITL suites for Gazebo Sim. 
\begin{itemize}
  \item Quadcopters: for PX4, the \textit{Holybro X500v2}, a simple carbon-fiber development platform with a 500mm frame and 10-inch propellers; for ArduPilot, the \textit{3DR Iris}, a legacy model with a 550mm frame, often used in academic research.
  \item QuadPlane VTOLs: for PX4, the \textit{Standard VTOL} model, based on a $\sim$2m wingspan airframe, with completely separate controls for multicopter (four electric lift motors) and fixed-wing flight (a rear pusher prop and elevons); for ArduPilot, the \textit{ALTI Transition QuadPlane}, a 3-meter wingspan commercial endurance VTOL also combining four vertical and one pusher motor. 
\end{itemize}

\subsubsection{On-board Sensors}
\label{sec:sensors}
We customize all four of these vehicle SDF models with a common suite of sensors to support advanced autonomy applications:

\begin{itemize} 
  \item RGB camera: a fixed front-facing EO sensor. It is configured to output a 320-by-240 pixel stream at 8Hz with a 100\textdegree\ \cameraready{horizontal} field-of-view (FOV), supporting visual servoing and YOLO-based object detection. \cameraready{This mimics the de-warped frame of a wide-angle CSI camera such as the IMX219-200.}
  \item 360\textdegree\ LiDAR: a 3D laser scanner modeled after the Livox Mid-360. It provides 360\textdegree\ horizontal and 60\textdegree\ vertical FOV (with 240 and 40 channels at 4Hz, respectively), generating point clouds for KISS-ICP odometry, obstacle avoidance, and SLAM.
\end{itemize}

These default sensor parameters are chosen to balance performance \cameraready{(see Section~\ref{sec:performance})} with realism. Nevertheless, they can be easily customized to simulate different or higher resolution sensors \cameraready{(e.g., increasing the sensor rate of the camera---for high-speed visual servoing---or that of the LiDAR---to more accurately model the 10Hz frequency of a real-life Livox Mid-360).}

\subsubsection{Worlds} We include a diverse set of 3D environments, each designed to validate perception-based autonomy in different scenarios.

\begin{itemize} 
  \item \textit{Empty}: a simple world with minimal computational overhead, ideal for unit testing and CI/CD pipelines. 
  \item \textit{Plain}: a flat environment featuring a high-resolution GIS ground texture for optical flow validation, populated with simple static obstacles (e.g., parked cars) to test basic object detection. 
  \item \textit{Mountain}: a 3D environment based on photogrammetry with a complex terrain topology, designed to develop and test terrain-following algorithms. 
  \item \textit{City}: a dense urban environment with high-rise structures, to test obstacle avoidance planners and occlusion handling in visual tracking. 
\end{itemize}

\begin{figure*}
    \centering
    \begin{tikzpicture} 
    \begin{axis}[
        width = 1.05\textwidth,
        height = 8.5cm,
        xmin=0,
        xmax=102,
        ymin=-20,
        ymax=120,   
        axis line style={draw=none},
        tick style={draw=none},
        xticklabel=\empty,
        yticklabel=\empty,
        clip=false,
]

\node[anchor=center,draw, align=center] (models) at (axis cs:10, 105) {\ttfamily \scriptsize aircraft\_models/};
    \node[anchor=center,draw, align=center] (worlds) at (axis cs:35, 105) {\ttfamily \scriptsize simulation\_worlds/};
    \node[anchor=center,draw, text width=2.05cm, align=center] (sitl) at (axis cs:10, 85) {\scriptsize 2 $\times$ {\ttfamily PX4} or \\[-1ex] {\ttfamily ArduPilot SITL}};
    \node[anchor=center,draw, text width=1cm, align=center] (gz) at (axis cs:37.5, 85) {\ttfamily \scriptsize Gazebo\\[-1ex] Sim};
\node[draw=blue, dashed, fit=(models) (worlds) (sitl) (gz), 
    label={[anchor=south west]north west:{\ttfamily \color{blue} \scriptsize simulation-container}}] (sim_box) {};
\node[draw, inner sep=12pt, fit=(sim_box), label={[anchor=south west]south west:{\ttfamily \bfseries \scriptsize amd64 host}}] (sim_host) {};

\node[anchor=center,draw, align=center] (gnd) at (axis cs:25, 20) {\ttfamily \scriptsize ground\_system};
    \node[anchor=center,draw, align=center] (qgc) at (axis cs:10, 42.5) {\ttfamily \scriptsize QGroundControl};
    \node[anchor=center,draw, align=center] (proxy) at (axis cs:25, 37.5) {\ttfamily \scriptsize mavlink-router};
    \node[anchor=center,draw, align=center] (zgnd) at (axis cs:10, 27.5) {\ttfamily \scriptsize zenoh-bridge};
\node[draw=blue, dashed, fit=(gnd) (qgc) (zgnd) (proxy),
    label={[anchor=south west]north west:{\ttfamily \color{blue} \scriptsize ground-container}}] (gnd_box) {};
\node[draw, inner sep=12pt, fit=(gnd_box), label={[anchor=south west]south west:{\ttfamily \bfseries \scriptsize amd64 host}}] (gnd_host) {};

\node[anchor=center,draw, align=center] (zair) at (axis cs:60, 60) {\ttfamily \scriptsize zenoh-bridge};
    \node[anchor=center,draw, align=center] (ss) at (axis cs:90, 60) {\ttfamily \scriptsize state\_sharing};
    \node[anchor=center,draw, text width=1.5cm, align=center] (dds) at (axis cs:60, 75) {\scriptsize {\ttfamily uxrce\_dds}\\[-1ex] or {\ttfamily MAVROS}};
    \node[anchor=center,draw, align=center] (kiss) at (axis cs:60, 95) {\ttfamily \scriptsize kiss\_icp};
    \node[anchor=center,draw, align=center] (ap) at (axis cs:90, 80) {\ttfamily \scriptsize autopilot\_interface};
    \node[anchor=center,draw, align=center] (yolo) at (axis cs:60, 110) {\ttfamily \scriptsize yolo\_py};
    \node[anchor=center,draw, align=center] (offboard) at (axis cs:75, 100) {\ttfamily \scriptsize offboard\_control};
    \node[anchor=center,draw, align=center] (mission) at (axis cs:90, 110) {\ttfamily \scriptsize mission};
\node[draw=blue, dashed, fit=(zair) (ss) (yolo) (kiss) (dds) (offboard) (mission) (ap),
    label={[anchor=south west]north west:{\ttfamily \color{blue} \scriptsize aircraft-container\_1}}] (air_box) {};
\node[draw, inner sep=12pt, fit=(air_box), label={[anchor=south west]south west:{\ttfamily \bfseries \scriptsize Jetson Orin NX}}] (jetson) {};

\node[anchor=center,draw, align=center] (zair2) at (axis cs:60, -25) {\ttfamily \scriptsize zenoh-bridge};
    \node[anchor=center,draw, align=center] (ss2) at (axis cs:90, -25) {\ttfamily \scriptsize state\_sharing};
    \node[anchor=center,draw, text width=1.5cm, align=center] (dds2) at (axis cs:60, -10) {\scriptsize {\ttfamily uxrce\_dds}\\[-1ex] or {\ttfamily MAVROS}};
    \node[anchor=center,draw, align=center] (kiss2) at (axis cs:60, 10) {\ttfamily \scriptsize kiss\_icp};
    \node[anchor=center,draw, align=center] (ap2) at (axis cs:90, -5) {\ttfamily \scriptsize autopilot\_interface};
    \node[anchor=center,draw, align=center] (yolo2) at (axis cs:60, 25) {\ttfamily \scriptsize yolo\_py};
    \node[anchor=center,draw, align=center] (offboard2) at (axis cs:75, 15) {\ttfamily \scriptsize offboard\_control};
    \node[anchor=center,draw, align=center] (mission2) at (axis cs:90, 25) {\ttfamily \scriptsize mission};
\node[draw=blue, dashed, fit=(zair2) (ss2) (yolo2) (kiss2) (dds2) (offboard2) (mission2) (ap2),
    label={[anchor=south west]north west:{\ttfamily \color{blue} \scriptsize aircraft-container\_2}}] (air_box2) {};
\node[draw, inner sep=12pt, fit=(air_box2), label={[anchor=south west]south west:{\ttfamily \bfseries \scriptsize Jetson Orin NX}}] (jetson2) {};

\draw[-latex] (models) to[out=330, in=150] (gz);
    \draw[-latex] (worlds) to[out=270, in=120] (gz);
    \draw[latex-latex] (sitl) to[out=0, in=180] node[pos=0.5, rounded corners=3pt,fill=black!10,text width=1.75cm,align=center,inner sep=2pt,] {\tiny {\ttfamily gz\_bridge} or\\[-3ex] {\ttfamily ardupilot\_gazebo}} (gz);

    \draw[-latex] (gnd) to[out=180, in=300] node[pos=0.4, rounded corners=3pt,fill=black!10,,align=center,inner sep=2pt,] {\ttfamily \tiny /tracks} (zgnd);
    \draw[-latex] (proxy) to[out=270, in=90] (gnd);
    \draw[latex-latex] (proxy) to[out=130, in=10] (qgc);

\draw[latex-latex] (dds) to[out=10, in=180] (ap);
    \draw[-latex] (dds) to[out=0, in=170] (ss);
    \draw[-latex] (yolo) to[out=10, in=100] node[pos=0.5, rounded corners=3pt,fill=black!10,,align=center,inner sep=2pt,] {\ttfamily \tiny /detections} (offboard);
\draw[-latex] (offboard) to[out=270, in=140] node[pos=0.33, rounded corners=3pt,fill=black!10,,align=center,inner sep=2pt,] {\ttfamily \tiny /reference} (ap);
    \draw[-latex] (mission) to[out=270, in=60] node[pos=0.4, rounded corners=3pt,fill=black!10,,text width=1.1cm,align=center,inner sep=2pt,] {\ttfamily \tiny ros2\\[-3ex] action/srv} (ap);
    \draw[latex-] (zair) to[out=0, in=180] node[pos=0.5, rounded corners=3pt,fill=black!10,,align=center,inner sep=2pt,] {\ttfamily \tiny /state\_drone\_1} (ss);

\draw[latex-latex] (dds2) to[out=10, in=180] (ap2);
    \draw[-latex] (dds2) to[out=0, in=170] (ss2);
    \draw[-latex] (yolo2) to[out=10, in=100] node[pos=0.5, rounded corners=3pt,fill=black!10,,align=center,inner sep=2pt,] {\ttfamily \tiny /detections} (offboard2);
\draw[-latex] (offboard2) to[out=270, in=140] node[pos=0.33, rounded corners=3pt,fill=black!10,,align=center,inner sep=2pt,] {\ttfamily \tiny /reference} (ap2);
    \draw[-latex] (mission2) to[out=270, in=60] node[pos=0.4, rounded corners=3pt,fill=black!10,,text width=1.1cm,align=center,inner sep=2pt,] {\ttfamily \tiny ros2\\[-3ex] action/srv} (ap2);
    \draw[latex-] (zair2) to[out=0, in=180] node[pos=0.5, rounded corners=3pt,fill=black!10,,align=center,inner sep=2pt,] {\ttfamily \tiny /state\_drone\_2} (ss2);

\node[anchor=center,draw, text width=1cm, align=center] (radio1) at (axis cs:5, -12.5) {\ttfamily \scriptsize radio\_0};
    \node[anchor=center,draw, text width=1cm, align=center] (radio2) at (axis cs:20, -12.5) {\ttfamily \scriptsize radio\_1};
    \node[anchor=center,draw, text width=1cm, align=center] (radio3) at (axis cs:12.5, -30) {\ttfamily \scriptsize radio\_2};
    \node[
font=\ttfamily\tiny,
        inner sep=2pt,
        yshift=-4pt,
        xshift=0pt,
    ] (manet) at (barycentric cs:radio1=1,radio2=1,radio3=1) {Mobile Ad-Hoc Network};
\node[draw=red, dashed, fit=(radio1) (radio2) (radio3), 
    label={[anchor=south west]north west:{\ttfamily \color{red} \bfseries \scriptsize AIR\_SUBNET}}] (airnet) {};
\draw[latex-latex,draw=red,line width=0.75pt,densely dashed] (zgnd) to[out=180, in=160] node[pos=0.35,text=red,fill=white,draw,,align=center,inner sep=2pt,] {\ttfamily \tiny TCP} (radio1);
    \draw[latex-latex,draw=red,line width=0.75pt,densely dashed] (zair) to[out=180, in=0] node[pos=0.85,text=red,fill=white,draw,,align=center,inner sep=2pt,] {\ttfamily \tiny TCP} (radio2);
    \draw[latex-latex,draw=red,line width=0.75pt,densely dashed] (zair2) to[out=190, in=-10] node[pos=0.5,text=red,fill=white,draw,,align=center,inner sep=2pt,] {\ttfamily \tiny TCP} (radio3);
\draw[latex-latex,draw] (radio1) to[out=250, in=170] (radio3);
    \draw[latex-latex,draw] (radio2) to[out=290, in=10] (radio3);
    \draw[latex-latex,draw] (radio1) to[out=10, in=170] (radio2);

\node[anchor=center,draw, text width=1cm, align=center] (router) at (axis cs:42.5, -10) {\ttfamily \scriptsize router};
\node[draw=teal, dashed, fit=(router), 
    label={[anchor=north west]south west:{\ttfamily \color{teal} \bfseries \scriptsize SIM\_SUBNET}}] (simnet) {};
\draw[latex-latex,draw=teal,line width=0.75pt] (sitl) to[out=-15, in=125,looseness=1.3] node[pos=0.31,text=teal, fill=white,draw,,align=center,inner sep=2pt,] {\ttfamily \tiny UDP} (router);
    \draw[latex-latex,draw=teal,line width=0.75pt] (sitl) to[out=-10, in=105, looseness=1.6] node[pos=0.28,text=teal, fill=white,draw,align=center,inner sep=2pt,] {\ttfamily \tiny MAVLink} (router);
    \draw[-latex,draw=teal,line width=0.75pt] (gz) to[out=310, in=85] node[pos=0.22, text=teal,fill=white,draw,align=center,inner sep=2pt,] {\ttfamily \tiny /lidar\_points} (router);
    \draw[-latex,draw=teal,line width=0.75pt] (gz) to[out=330, in=65] node[pos=0.12,text=teal, fill=white,draw,align=center,inner sep=2pt,] {\ttfamily \tiny gz\_gst\_bridge} (router);
\draw[latex-latex,draw=teal,line width=0.75pt] (router) to[out=150, in=-20] (proxy);
\draw[latex-latex,draw=teal,line width=0.75pt] (router) to[out=20, in=180,looseness=0.4] (dds);
    \draw[-latex,draw=teal,line width=0.75pt] (router) to[out=28, in=180,looseness=0.5] (kiss);
    \draw[-latex,draw=teal,line width=0.75pt] (router) to[out=40, in=180,looseness=0.5] (yolo);
\draw[latex-latex,draw=teal,line width=0.75pt] (router) to[out=-20, in=180,looseness=0.5] (dds2);
    \draw[-latex,draw=teal,line width=0.75pt] (router) to[out=-10, in=180,looseness=0.5] (kiss2);
    \draw[-latex,draw=teal,line width=0.75pt] (router) to[out=0, in=180,looseness=0.5] (yolo2);

\path (gz) to[out=310, in=85] 
    node[
        pos=0.22, 
        text=teal,
        fill=white,
        draw=teal,
        align=center,
        inner sep=2pt
    ] {\ttfamily \tiny /lidar\_points} 
    (router);

    \end{axis}
\end{tikzpicture}     \caption{
    Block diagram of the hardware-in-the-loop simulation architecture for two vehicles, with the \emph{simulation-image} and \emph{ground-image} containers running on separate \texttt{amd64} hosts, and two \emph{aircraft-image} containers on two dedicated NVIDIA Jetson Orin.
    The \texttt{SIM\_SUBNET} is created with a router over a wired Ethernet network while the \texttt{AIR\_SUBNET} is created over a wireless mobile ad-hoc network.
    }
    \label{fig:hitl}
    \vspace{-1.5em}
\end{figure*}
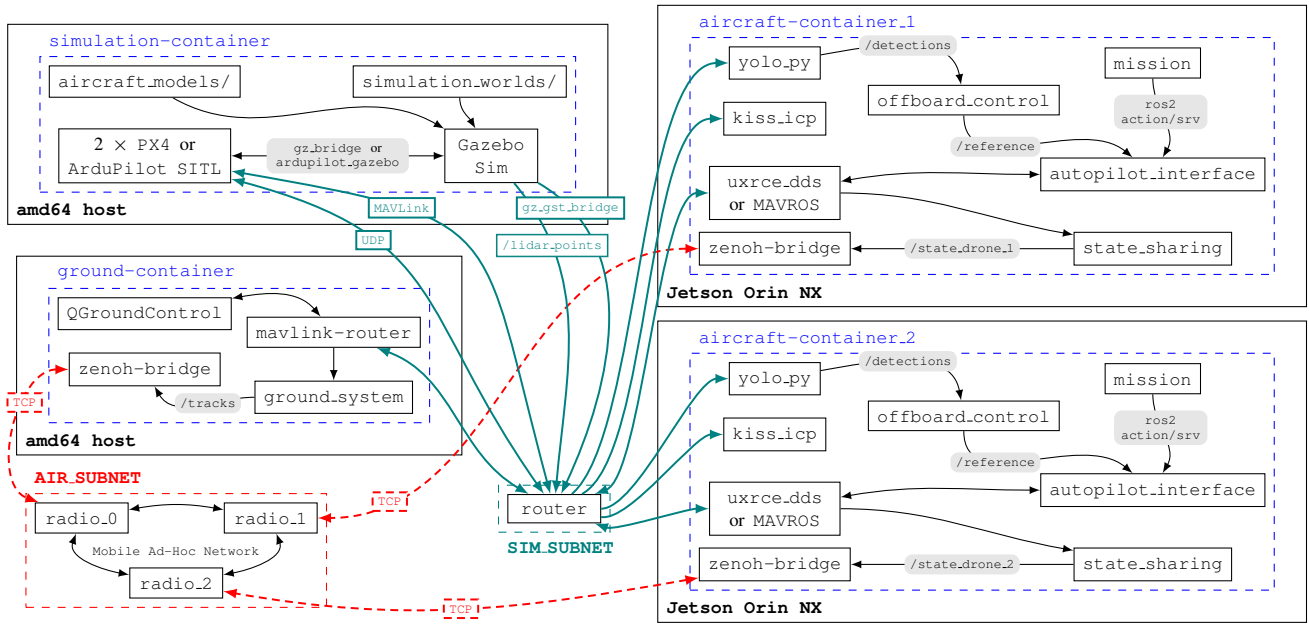

\subsection{Dockerization}

The \texttt{aerial-autonomy-stack} is distributed as a Dockerized application to maximize its portability.
We tested it on Ubuntu 22.04 LTS, 24.04 LTS, and Windows 11 Pro/Enterprise via WSL2. 
Two critical requirements are \emph{(i)} the NVIDIA graphics driver and \emph{(ii)} the NVIDIA Container Toolkit, which enable containerized processes to access the host GPU for OpenGL rendering and CUDA inference.

The stack is architected around three container images: 
\begin{itemize} 
	\item the \emph{simulation-image} contains Gazebo Sim, the compiled PX4/ArduPilot SITL binaries, and the complete library of aircraft and world SDF assets; it serves as the digital twin of the physical environment;
	\item the \emph{ground-image} hosts QGroundControl, for operator oversight, alongside the MAVLink and Zenoh bridges; 
	\item the \emph{aircraft-image} contains the ROS2 autonomy stack, the autopilot interface middleware (uXRCE-DDS or MAVROS), and the perception pipelines; crucially, it is created from the same Dockerfile for both simulation and deployment, minimizing the sim2real gap.
\end{itemize}

\subsubsection{Multi-target Compilation}

Supporting both simulation and flight hardware requires multi-architecture builds to enable a ``write once, deploy everywhere'' philosophy.

\begin{itemize} 
	\item \texttt{amd64} images: we start from a base image supporting the \texttt{x86\_64} architecture for all three images (simulation, ground, aircraft) to enable high-performance simulation on developer workstations and CI servers. 
	\item \texttt{arm64}/Jetson images: we also include an \texttt{aarch64} base image specifically for the \emph{aircraft-image}. This supports compilation and deployment on NVIDIA Jetson\footnote{\url{https://developer.nvidia.com/embedded/jetpack}} computers (JetPack 6/L4T 36) without any code change.
\end{itemize}

\paragraph{YOLO on CUDA/TensorRT} We also adopt a dual-architecture strategy for inference. On the \texttt{amd64} images for simulation, we utilize the ONNX CUDA runtime, while \texttt{arm64} images for deployment on Jetson leverage the ONNX TensorRT runtime. For the latter, we quantize the model weights to FP16, \cameraready{achieving inference at over 50Hz with the IMX219-200 CSI camera (acquiring 1280-by-720 pixel frames at 60Hz) on Orin NX using the nano YOLO model with input size of 640}.

\subsubsection{Networking}

A major differentiating aspect of the \texttt{aerial-autonomy-stack} is its ability to capture the networking architecture of a multi-robot system from the early stages of development. Using Docker, we isolate containers on two virtual networks, forcing their components to communicate via standard IP protocols.

The ``internal'' \texttt{SIM\_SUBNET} mimics the physical cabling inside a drone's frame.
\begin{itemize} 
	\item The UDP connection between autopilot SITL and DDS/MAVROS  simulates the high-speed Ethernet link between a Pixhawk 6X flight controller and the Jetson. 
	\item The UDP GStreamer stream simulates the CSI camera interface, piping raw video frames from the Gazebo sensor to the perception node. 
	\item The ROS2 LiDAR bridge simulates the Ethernet connection of a physical LiDAR. 
	\item The UDP MAVLink connection provides telemetry streams (split by mavproxy) to QGroundControl and the \emph{ground-image} container. 
\end{itemize}

The ``external'' \texttt{AIR\_SUBNET} mimics the IP radio link between multiple drones and a ground station computer. 
\begin{itemize} 
	\item This is the network hosting the Zenoh-ROS2 bridge for inter-robot state-sharing, optimized to handle discovery and serialization over lossy Wi-Fi or mesh links.
\end{itemize}

 \section{Software-in-the-loop Capabilities} 
\label{sec:sitl}

This work expands upon traditional SITL workflows by shifting the focus from a component-based approach to a system-holistic one. 
By encapsulating the entire development environment, \texttt{aerial-autonomy-stack} allows researchers to validate complex behaviors---involving perception and control---safely and efficiently before committing to expensive flight hours.

\subsection{Networked Multi-robot Simulation}

Developing and testing multi-agent systems typically requires a fleet of physical machines or a complex cluster. 
Our stack allows to simulate \emph{networked} swarms of quadcopters and VTOLs on a single developer's laptop.
This consolidation is typically challenging because of namespace collisions and resource contention. 

\textbf{Implementation:} We leverage Docker's network isolation to assign each simulated vehicle its own virtual IP address and unique \texttt{ROS\_DOMAIN\_ID}. 
This architecture ensures that agents communicate only through defined channels, accurately mirroring the constraints of a physical ad-hoc network while executing on a single host.

\subsection{Faster-than-real-time Simulation}

We provide native support for FTRT simulation, incorporating full perception pipelines and pause/resume functionality, essential for reinforcement learning applications.

Accelerating simulation is not merely a matter of increasing frame rates; it requires strict temporal synchronization across the entire stack. 
If the simulation runs faster than the wall-clock, the autopilot and perception nodes must be decoupled from the system time and driven exclusively by the synthetic clock. 
Failure to enforce this lockstep would result in watchdog timeouts, where estimators reject sensor data, leading to catastrophic instability.

\textbf{Implementation:} To address this, we configure the \texttt{ros\_gz\_bridge} to broadcast the simulation time to the \texttt{/clock} topic. All ROS2 nodes are configured to subscribe to this source.
When scaling to multiple agents, this requires careful orchestration across ROS domains to ensure all agents synchronize to the master simulation clock.

Achieving FTRT with high-fidelity perception is computationally demanding. 
CPU-based rendering constitutes a significant bottleneck. 
For rendering, we utilize the NVIDIA Container Toolkit to let host GPU passthrough into the Docker containers, allowing Gazebo's Ogre2 engine to render camera and LiDAR frames at rates far exceeding real-time requirements. 
Similarly, CUDA-accelerated YOLO models leverage the mapped GPU to perform inference on every simulated frame without stalling the physics loop. 

\subsection{End-to-end, Vertically Integrated CI/CD}

Finally, the \texttt{aerial-autonomy-stack} allows the validation of an end-to-end system---from perception to actuation---within a unified CI/CD pipeline.

\textbf{Implementation:} Traditional development often relies on unit tests that mock external dependencies, failing to capture integration errors (such as version mismatches). 
We mitigate this by using a single Dockerfile to create the same \emph{aircraft-image} for both \texttt{amd64} simulation and \texttt{arm64} deployment. 
Vertical integration ensures that a successful pass of the CI/CD pipeline guarantees the compatibility of the entire software stack, including middleware and libraries, significantly reducing the ``it works on my machine'' syndrome during field trials.
 \section{Hardware-in-the-loop Capabilities} 
\label{sec:hitl}

Hardware-in-the-loop simulation is an essential validation step for safety-critical aerospace systems, narrowing the gap between pure simulation and flight tests.
Traditionally, HITL has focused on the validation of the microcontroller-based flight controller unit (FCU). 
However, with modern SITL simulation---where the exact firmware binary executes against the simulator with synchronized clocks---the role of FCU-level HITL is partially diminished. 

In perception-based autonomy, a potentially more critical ``architectural sim2real gap'' lies not in the flight controller, but in the companion computer.
The disparity in architecture---typically \texttt{x86\_64} for development and \texttt{arm64} for deployment---combined with vast differences in GPUs, thermal budgets, and memory bandwidth, often results in severe performance degradation when moving from a developer's workstation to the deployment edge device. 
Furthermore, accurate validation of multi-robot behaviors requires testing the actual physical communication layer, which simulations cannot perfectly model.

To address these blind spots, our stack supports two distinct levels of HITL validation (Figure~\ref{fig:hitl}).

\subsection{Jetson-in-the-loop Validation}

The primary goal is to validate the computational viability of the perception and control stack on the target hardware. This validation ensures that deep learning models fit within the constrained GPU memory of the edge device and that inference latency remains acceptable under thermal load, preventing unexpected frequency throttling during flight.

\textbf{Implementation:} We leverage the Dockerized architecture to decouple the simulation and the autonomy components of the stack.
The \emph{simulation-image} container runs on a high-performance desktop workstation, handling physics and rendering.
The \emph{aircraft-image} container is deployed directly onto the physical NVIDIA Jetson Orin modules.
These are connected via two Gigabit Ethernet interfaces (one for \texttt{SIM\_SUBNET} and one for \texttt{AIR\_SUBNET})
The workstation streams simulated camera feeds and autopilot telemetry to the Jetson, mimicking physical sensors, while the Jetsons perform TensorRT inference, compute control actions, and send them back to the simulator via the autopilot interface.

\subsection{Comms-in-the-loop Validation}

This mode aims to stress-test the multi-agent coordination logic against the realities of RF communication---such as bandwidth saturation, packet loss, and latency jitter---which are difficult to model synthetically.

\textbf{Implementation:} Expanding on the Jetson-in-the-loop setup, we replace the Ethernet link for the \texttt{AIR\_SUBNET} with the actual wireless hardware (e.g., Mesh radios) intended for the mission. 
The simulation computer and the Jetsons remain connected through the \texttt{SIM\_SUBNET} for the synthetic high-bandwidth sensor data, while the inter-vehicle information sharing on the \texttt{AIR\_SUBNET} moves exclusively over the air. 
This exposes the system to real-world interference and connection drops, allowing developers to tune Quality of Service (QoS) policies and robustness logic even before the field trials.
 \begin{figure*}
    \centering
\begin{lstlisting}[
        firstnumber=1,
        numbers=none,
        language=Bash, 
        label = {alg:api},
        ]
# 1. Takeoff action (PX4/ArduPilot quads and VTOLs)
ros2 action send_goal /Drone${DRONE_ID}/takeoff_action autopilot_interface_msgs/action/Takeoff 
    '{takeoff_altitude: 40.0, 
    vtol_transition_heading: 0.0, vtol_loiter_nord: 0.0, vtol_loiter_east: 0.0, vtol_loiter_alt: 100.0}'
# 2. Orbit action (PX4/ArduPilot quads and VTOLs)
ros2 action send_goal /Drone${DRONE_ID}/orbit_action autopilot_interface_msgs/action/Orbit 
    '{east: 500.0, north: 0.0, altitude: 150.0, radius: 200.0}'
# 3. Land (at home) action (PX4/ArduPilot quads and VTOLs)
ros2 action send_goal /Drone${DRONE_ID}/land_action autopilot_interface_msgs/action/Land 
    '{landing_altitude: 60.0, vtol_transition_heading: 60.0}'
# 4. Offboard action (PX4 quads and VTOLs, ArduPilot quads)
# - PX4 offboard_setpoint_type: attitude = 0, rates = 1, trajectory = 2
# - ArduPilot quads offboard_setpoint_type: velocity = 3, acceleration = 4
ros2 action send_goal /Drone${DRONE_ID}/offboard_action autopilot_interface_msgs/action/Offboard 
    '{offboard_setpoint_type: 1, max_duration_sec: 5.0}'
\end{lstlisting}
    \vspace{-0.5em} 
    \caption{
     Command-line interface for the four high-level ROS2 Actions (\textbf{Takeoff}, \textbf{Orbit}, \textbf{Land}, \textbf{Offboard}) provided by the autopilot-agnostic interface.
    }
    \label{fig:api}
    \vspace{-1.5em}
\end{figure*}

\section{Autopilot-agnostic ROS2 Actions}
\label{sec:ap-interface}

Advanced autonomy requires communication paradigms that support long-running behaviors with continuous feedback---a capability often lacking in basic publisher/subscriber or service/request models. ROS2 Actions address this gap by allowing a client to initiate a goal (e.g., Takeoff) and receive periodic updates on progress (e.g., the current altitude) until the final success or failure state.

PX4's internal uORB messaging system is inherently publisher/subscriber, and its DDS bridge exposes these topics directly to ROS2, requiring a developer to manage state machines manually. Conversely, MAVROS (which our stack uses for ArduPilot) was originally developed on ROS1 and relies heavily on blocking service/request calls, which can stall the execution pipeline.

In \texttt{aerial-autonomy-stack}, we introduce a modern abstraction layer that sits atop these autopilot-specific implementations. The \texttt{autopilot\_interface} package provides a unified set of ROS2 Actions, allowing developers to write control logic that is agnostic to the underlying (PX4 or ArduPilot) firmware. Its interface exposes the following high-level behaviors (Figure~\ref{fig:api}):
\begin{itemize} 
	\item \textbf{Takeoff} initiates the arming sequence and commands an ascent to a target altitude. For VTOLs, it includes a transition heading and a parking loiter.
	\item \textbf{Orbit} commands the vehicle to loiter around a specific location at a defined distance and altitude. For multicopters, it includes yaw control towards the center.
	\item \textbf{Land} executes a landing at home sequence, managing the descent and, for VTOLs, the transition heading.
	\item \textbf{Offboard} is a flexible action for low-level control, allowing the companion computer to bypass the autopilot's internal navigator and drive the control loops directly. It supports multiple setpoint modes: 
	\begin{itemize} 
		\item \textit{Trajectory (PX4)}: specific coordinates and velocities for precise waypoint navigation. 
		\item \textit{Velocity (ArduPilot)}: 3D velocity setpoints, ideal for visual servoing and target following. 
		\item \textit{Acceleration (ArduPilot)}: 3D acceleration setpoints, useful for trajectory smoothing. 
		\item \textit{Attitude (PX4)}: vehicle orientation and collective thrust for low-level control. 
		\item \textit{Rates (PX4)}: direct body-rate control for the most aggressive maneuvers. 
	\end{itemize} 
\end{itemize}

Finally, for multicopter configurations, we also expose a \textbf{Reposition} service. This allows for simple go-to commands that delegate path planning to the autopilot's internal navigator.
This serves as a lightweight alternative to \textbf{Offboard} control for simple point-to-point navigation where the explicit enforcement of lower-level constraints is not required.
 \section{Performance Results}
\label{sec:performance}

\begin{table}
    \centering
    \caption{
    Simulation Clock Speed-ups on a Mobile Workstation\\
    (Intel i9-13 with 64GB RAM and 12GB NVIDIA RTX3500).
    }
    \scalebox{0.84}{

\begin{tabular}{ 
c
c
c
c
c
c
c
} 

\toprule

\multirow{2}{*}{\textbf{Autopilot}} &
\textbf{Sim.} &
\textbf{Num.} & 
\textbf{Neither} &
\textbf{Camera} &
\textbf{LiDAR} &
\textbf{Camera \&}
\tabularnewline 
& 
\textbf{Instances} &
\textbf{Drones} &
\textbf{Sensor} &
\textbf{Only} &
\textbf{Only} &
\textbf{LiDAR}
\tabularnewline 

\cmidrule(lr){1-7}

\multirow{7}{*}{PX4} 
&
\multirow{4}{*}{1}
&
1
&
10.56$\times$
&
10.93$\times$
&
10.89$\times$
&
10.49$\times$
\tabularnewline

&

&
2
&
9.39$\times$
&
6.88$\times$
&
8.45$\times$
&
6.05$\times$
\tabularnewline

&

&
4
&
4.97$\times$
&
3.05$\times$
&
4.26$\times$
&
2.76$\times$
\tabularnewline

&

&
6 
&
3.19$\times$
&
1.93$\times$
&
2.65$\times$
&
1.74$\times$
\tabularnewline 

\cmidrule(lr){2-3}
\cmidrule(lr){4-7}

&
\multirow{3}{*}{2} 
&
1 ea.
&
\bfseries 26.05$\times$
&
\bfseries 20.01$\times$
&
\bfseries 21.73$\times$
&
\bfseries 17.53$\times$
\tabularnewline

&

&
2 ea.
&
15.71$\times$
&
10.72$\times$
&
12.22$\times$
&
9.01$\times$
\tabularnewline

&

&
3 ea.
&
14.72$\times$
&
9.15$\times$
&
11.36$\times$
&
7.02$\times$
\tabularnewline

\cmidrule(lr){1-7}

\multirow{7}{*}{ArduPilot} 
&
\multirow{4}{*}{1}
&
1
&
8.35$\times$
&
8.76$\times$
&
9.85$\times$
&
8.39$\times$
\tabularnewline

&

&
2
&
6.84$\times$
&
4.93$\times$
&
6.47$\times$
&
4.40$\times$
\tabularnewline

&

&
4
&
3.44$\times$
&
2.25$\times$
&
3.00$\times$
&
2.09$\times$
\tabularnewline

&

&
6 
&
1.82$\times$
&
1.34$\times$
&
1.80$\times$
&
1.22$\times$
\tabularnewline 

\cmidrule(lr){2-3}
\cmidrule(lr){4-7}

&
\multirow{3}{*}{2} 
&
1 ea.
&
15.06$\times$
&
13.05$\times$
&
14.70$\times$
&
12.24$\times$
\tabularnewline

&

&
2 ea.
&
9.47$\times$
&
7.69$\times$
&
7.83$\times$
&
6.96$\times$
\tabularnewline

&

&
3 ea. 
&
7.50$\times$
&
5.15$\times$
&
6.69$\times$
&
5.34$\times$
\tabularnewline

\bottomrule
\end{tabular}
}
     \label{tab:performance}
\end{table}

Table~\ref{tab:performance} summarizes the faster-than-real-time throughput achieved by the \texttt{aerial-autonomy-stack} on a mobile workstation with an 13th Gen Intel Core i9 CPU, 64GB of RAM, and a 12GB NVIDIA RTX3500 Ada Generation GPU. 

Column ``Sim. Instances'' denotes whether the workload was executed on a single instance of the stack or distributed across two parallel instances (splitting the drone fleet evenly).

The ODE physics timestep for the Gazebo environment is maintained at the default 4ms (250Hz) for PX4-based simulations, while for ArduPilot, we increase the default 1ms to 2ms (500Hz) to balance fidelity with performance.
On our testing hardware, the physics engine can sustain a raw Real-Time Factor (RTF) of approximately 15--20$\times$, however, we cap it to 15$\times$ to avoid jitter and ensure autopilot SITL stability.
Thus, the theoretical upper bound for faster-than-real-time sample generations is $\leq$15$\times$ for the single-instance experiments and $\leq$30$\times$ for the dual-instance configuration.

Each cell in Table~\ref{tab:performance} is the average of 10 runs, each 250 simulated seconds long.
We evaluate the faster-than-real-time speed-ups when simulating 1, 2, 4, and 6 quadcopters simultaneously.
For each drone, we vary the computational load by including either both, each one, or neither of two simulated sensors: \emph{(i)} a 320-by-240 pixels, 8 FPS, RGB camera, processed by Ultralytics YOLOv8n~\cite{yolo-review} at native resolution; and \emph{(ii)} a 4Hz LiDAR with 360\textdegree\ (240 samples) horizontal FOV and 60\textdegree\ (40 samples) vertical FOV processed by KISS-ICP~\cite{kiss-icp} odometry. 
\cameraready{Note that these sensor configurations are simplifications of the real-life sensors (see Section~\ref{sec:sensors}) and they have been selected to balance simulation performance and realism. Nonetheless, all sensor parameters can be easily modified in software.}

Our results show that \texttt{aerial-autonomy-stack} consistently achieves above real-time performance across all tested scenarios.
Furthermore, in optimized configurations, it attains an effective aggregate speed-up in the order of $\sim$20$\times$.
\cameraready{Multi-drone scenarios are slower but still run faster than real-time, e.g., between $\sim$3$\times$ and $\sim$10$\times$ for 4 PX4 drones.}
Crucially, we observe that the system is primarily compute-bound by CPU performance---saturated by physics updates and inter-process communication---rather than GPU inference capability, highlighting the efficiency of our hardware-accelerated perception stack.
 \section{Conclusions}
\label{sec:conclusions}

We introduced \texttt{aerial-autonomy-stack}, an open-source, vertically integrated framework to help overcome the fragmentation of modern aerial robotics tooling. 
Through meticulous hardware-software co-design, we fulfill the need for a unified perception-to-control workflow, enabling seamless transitions from simulation to deployment. 
We successfully demonstrated that our stack supports faster-than-real-time simulation on consumer GPUs.
Furthermore, using containerization, we enabled networked multi-agent simulation and frictionless deployment to Jetson edge devices.

Current limitations include vehicle support restricted to multirotor and quad-plane VTOL configurations, and the lack of offboard multi-sensor fusion. 
Although reinforcement learning APIs are part of the \texttt{aerial-autonomy-stack} package, they were excluded from this article to maintain a sharper focus on the system architecture and development/deployment cycle. 
Future work will focus on SLAM integration, end-to-end learning, and outdoor field validation.
 \section{Acknowledgments}
\label{sec:acknowledgments}
This work is supported by National Research Council Canada (NRC) and Defence Research and Development Canada (DRDC), partially, by its Canadian Saftey and Security program (DRDC CSS). The Canadian Safety and Security Program is a federally-funded program to strengthen Canada's ability to anticipate, prevent/mitigate, prepare for, respond to, and recover from natural disasters, serious accidents, crime and terrorism through the convergence of science and technology with policy, operations and intelligence.  
\balance

\balance

\end{document}